# A Proposal for Word Sense Disambiguation using Conceptual Distance

**Eneko Agirre.***
Lengoaia eta Sistema Informatikoak saila.
Euskal Herriko Unibertsitatea.
p.k. 649, 20080 Donostia. Spain. jibagbee@si.ehu.es

**German Rigau.****
Departament de Llenguatges i Sistemes Informàtics.
Universitat Politècnica de Catalunya.
Pau Gargallo 5, 08028 Barcelona. Spain. g.rigau@lsi.upc.es

## Abstract.

This paper presents a method for the resolution of lexical ambiguity and its automatic evaluation over the Brown Corpus. The method relies on the use of the wide-coverage noun taxonomy of WordNet and the notion of conceptual distance among concepts, captured by a Conceptual Density formula developed for this purpose. This fully automatic method requires no hand coding of lexical entries, hand tagging of text nor any kind of training process. The results of the experiment have been automatically evaluated against SemCor, the sense-tagged version of the Brown Corpus.

## 1 Introduction

Word sense disambiguation is a long-standing problem in Computational Linguistics. Much of recent work in lexical ambiguity resolution offers the prospect that a disambiguation system might be able to receive as input unrestricted text and tag each word with the most likely sense with fairly reasonable accuracy and efficiency. The most extended approach is to attempt to use the context of the word to be disambiguated together with information about each of its word senses to solve this problem.

Several interesting experiments have been performed in recent years using preexisting lexical knowledge resources. (Cowie et al. 92) describe a method for lexical disambiguation of text using the definitions in the machine-readable version of the LDOCE dictionary as in the method described in (Lesk 86), but using simulated annealing for efficiency reasons. (Yarowsky 92) combines the use of the Grolier encyclopaedia as a training corpus with the categories of the Roget's International Thesaurus to create a statistical model for the word sense disambiguation problem with excellent results. (Wilks et al. 93) perform several interesting statistical disambiguation experiments using coocurrence data collected from LDOCE. (Sussna 93), (Voorhees 93), (Richarson et al. 94) define a disambiguation programs based in WordNet with the goal of improving precision and coverage during document indexing.

Although each of these techniques looks somewhat promising for disambiguation, either they have been only applied to a small number of words, a few sentences or not in a public domain corpus. For this reason we have tried to disambiguate all the nouns from real texts in the public domain sense tagged version of the Brown corpus (Francis & Kucera 67), (Miller et al. 93), also called Semantic Concordance or Semcor for short. We also use a public domain lexical knowledge source, WordNet (Miller 90). The advantage of this approach is clear, as Semcor provides an appropriate environment for testing our procedures in a fully automatic way.

*\* Eneko Agirre was supported by a grant from the Basque Government.*
*\*\* German Rigau was supported by a grant from the Ministerio de Educación y Ciencia.*

This paper presents a general automatic decision procedure for lexical ambiguity resolution based on a formula of the conceptual distance among concepts: Conceptual Density. The system needs to know how words are clustered in semantic classes, and how semantic classes are hierarchically organised. For this purpose, we have used a broad semantic taxonomy for English, WordNet. Given a piece of text from the Brown Corpus, our system tries to resolve the lexical ambiguity of nouns by finding the combination of senses from a set of contiguous nouns that maximises the total Conceptual Density among senses.

Even if this technique is presented as stand-alone, it is our belief, following the ideas of (McRoy 92) that full-fledged lexical ambiguity resolution should combine several information sources. Conceptual Density might be only one evidence of the plausibility of a certain word sense.

Following this introduction, section 2 presents the semantic knowledge sources used by the system. Section 3 is devoted to the definition of Conceptual Density. Section 4 shows the disambiguation algorithm used in the experiment. In section 5, we explain and evaluate the performed experiment. In section 6, we present further work and finally in the last section some conclusions are drawn.

## 2 WordNet and the Semantic Concordance

Sense is not a well defined concept and often has subtle distinctions in topic, register, dialect, collocation, part of speech, etc. For the purpose of this study, we take as the senses of a word those ones present in WordNet 1.4. WordNet is an on-line lexicon based on psycholinguistic theories (Miller 90). It comprises nouns, verbs, adjectives and adverbs, organised in terms of their meanings around semantic relations, which include among others, synonymy and antonymy, hypernymy and hyponymy, meronymy and holonymy. Lexicalised concepts, represented as sets of synonyms called synsets, are the basic elements of WordNet. The senses of a word are represented by synsets, one for each word sense. The version used in this work, WordNet 1.4, contains 83,800 words, 63,300 synsets (word senses) and 87,600 links between concepts.

The nominal part of WordNet can be viewed as a tangled hierarchy of hypo/hypernymy relations. Nominal relations include also three kinds of meronymic relations, which can be paraphrased as member-of, made-of and component-part-of.

SemCor (Miller et al. 93) is a corpus where a single part of speech tag and a single word sense tag (which corresponds to a WordNet synset) have been included for all open-class words. SemCor is a subset taken from the Brown Corpus (Francis & Kucera, 67) which comprises approximately 250,000 words out of a total of 1 million words. The coverage in WordNet of the senses for open-class words in SemCor reaches 96% according to the authors. The tagging was done manually, and the error rate measured by the authors is around 10% for polysemous words.

## 3 Conceptual Density and Word Sense Disambiguation

A measure of the relatedness among concepts can be a valuable prediction knowledge source to several decisions in Natural Language Processing. For example, the relatedness of a certain word-sense to the context allows us to select that sense over the others, and actually disambiguate the word. Relatedness can be measured by a fine-grained conceptual distance (Miller & Teibel, 91) among concepts in a hierarchical semantic net such as WordNet. This measure would allow to discover reliably the lexical cohesion of a given set of words in English.

Conceptual distance tries to provide a basis for determining closeness in meaning among words, taking as reference a structured hierarchical net. Conceptual distance between two concepts is defined in (Rada et al. 89) as the length of the shortest path that connects the concepts in a hierarchical semantic net. In a similar approach, (Sussna 93) employs the notion of conceptual distance between network nodes in order to improve precision during document indexing. Following these ideas, (Agirre et al. 94) describes a new conceptual distance formula for the automatic spelling correction problem and (Rigau 94), using this conceptual distance formula, presents a methodology to enrich dictionary senses with semantic tags extracted from WordNet.

The measure of conceptual distance among concepts we are looking for should be sensitive to:

• the length of the shortest path that connects the concepts involved.

- the depth in the hierarchy: concepts in a deeper part of the hierarchy should be ranked closer.
- the density of concepts in the hierarchy: concepts in a dense part of the hierarchy are relatively closer than those in a more sparse region.
- the measure should be independent of the number of concepts we are measuring.

We have experimented with several formulas that follow the four criteria presented above. Currently, we are working with the Conceptual Density formula, which compares areas of subhierarchies.

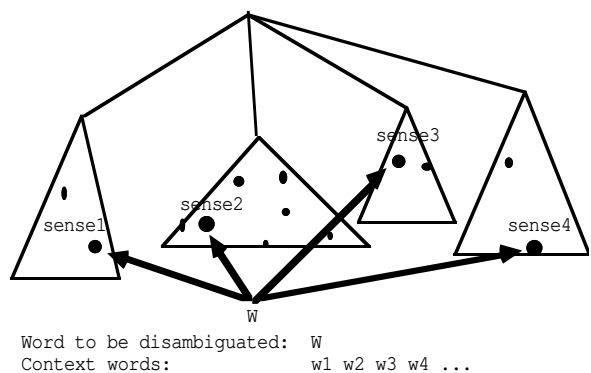

Word to be disambiguated:   W
Context words:              w1 w2 w3 w4 ...

Figure 1: senses of a word in WordNet

As an example of how Conceptual Density can help to disambiguate a word, in figure 1 the word W has four senses and several context words. Each sense of the words belongs to a subhierachy of WordNet. The dots in the subhierarchies represent the senses of either the word to be disambiguated (W) or the words in the context. Conceptual Density will yield the highest density for the subhierarchy containing more senses of those, relative to the total amount of senses in the subhierarchy. The sense of W contained in the subhierarchy with highest Conceptual Density will be chosen as the sense disambiguating W in the given context. In figure 1, sense2 would be chosen.

Given a concept $c$, at the top of a subhierarchy, and given $nhyp$ and $h$ (mean number of hyponyms per node and height of the subhierarchy, respectively), the Conceptual Density for $c$ when its subhierarchy contains a number $m$ (marks) of senses of the words to disambiguate is given by the formula below:

$$CD(c,m) = \frac{\sum_{i=0}^{m-1} nhyp^i}{\sum_{i=0}^{h-1} nhyp^i} \qquad (1)$$

The numerator expresses the expected area for a subhierarchy containing $m$ marks (senses of the words to be disambiguated), while the divisor is the actual area, that is, the formula gives the ratio between weighted marks below $c$ and the number of descendant senses of concept $c$. In this way, formula 1 captures the relation between the weighted marks in the subhierarchy and the total area of the subhierarchy below $c$. The weight given to the marks tries to express that the height and the number of marks should be proportional.

$nhyp$ is computed for each concept in WordNet in such a way as to satisfy equation 2, which expresses the relation among height, averaged number of hyponyms of each sense and total number of senses in a subhierarchy if it were homogeneous and regular:

$$descendants_c = \sum_{i=0}^{h-1} nhyp^i \qquad (2)$$

Thus, if we had a concept $c$ with a subhierarchy of height 5 and 31 descendants, equation 2 will hold that $nhyp$ is 2 for $c$.

Conceptual Density weights the number of senses of the words to be disambiguated in order to make density equal to 1 when the number $m$ of senses below $c$ is equal to the height of the hierarchy $h$, to make density smaller than 1 if $m$ is smaller than $h$ and to make density bigger than 1 whenever $m$ is bigger than $h$. The density can be kept constant for different $m$-s provided a certain proportion between the number of marks $m$ and the height $h$ of the subhierarchy is maintained. Both hierarchies **A** and **B** in figure 2, for instance, have Conceptual Density 1.

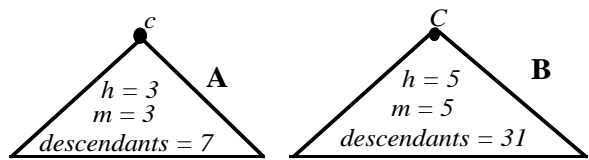

Figure 2: two hierarchies with CD = 1[1].

---

[1]*From formulas 1 and 2 we have:*

$descendants(c) = 7 = \sum_{i=0}^{3-1} nhyp^i \Rightarrow nhyp = 2 \Rightarrow CD(c,3) = \sum_{i=0}^{3-1} 2^i \Big/ 7 = 7/7 = 1$

$descendants(c) = 31 = \sum_{i=0}^{5-1} nhyp^i \Rightarrow nhyp = 2 \Rightarrow CD(c,5) = \sum_{i=0}^{5-1} 2^i \Big/ 31 = 31/31 = 1$

In order to tune the Conceptual Density formula, we have made several experiments adding two parameters, α and β. The a parameter modifies the strength of the exponential *i* in the numerator because *h* ranges between 1 and 16 (the maximum number of levels in WordNet) while *m* between 1 and the total number of senses in WordNet. Adding a constant b to *nhyp*, we tried to discover the role of the averaged number of hyponyms per concept. Formula 3 shows the resulting formula.

$$CD(c, m) = \frac{\sum\limits_{i=0}^{m-1} (nhyp + \beta)^{i^{\alpha}}}{\text{descendants}_c} \qquad \textbf{(3)}$$

After an extended number of runs which were automatically checked, the results showed that β does not affect the behaviour of the formula, a strong indication that this formula is not sensitive to constant variations in the number of hyponyms. On the contrary, different values of α affect the performance consistently, yielding the best results in those experiments with α near 0.20. The actual formula which was used in the experiments was thus the following:

$$CD(c, m) = \frac{\sum\limits_{i=0}^{m-1} nhyp^{i^{0.20}}}{\text{descendants}_c} \qquad \textbf{(4)}$$

## 4 The Disambiguation Algorithm Using Conceptual Density

Given a window size, the program moves the window one word at a time from the beginning of the document towards its end, disambiguating in each step the word in the middle of the window and considering the other words in the window as context.

The algorithm to disambiguate a given word w in the middle of a window of words W roughly proceeds as follows. First, the algorithm represents in a lattice the nouns present in the window, their senses and hypernyms (step 1). Then, the program computes the Conceptual Density of each concept in WordNet according to the senses it contains in its subhierarchy (step 2). It selects the concept c with highest density (step 3) and selects the senses below it as the correct senses for the respective words (step 4). If a word from W:

• has a single sense under c, it has already been disambiguated.
• has not such a sense, it is still ambiguous.
• has more than one such senses, we can eliminate all the other senses of w, but have not yet completely disambiguated w.

The algorithm proceeds then to compute the density for the remaining senses in the lattice, and continues to disambiguate words in W (back to steps 2, 3 and 4). When no further disambiguation is possible, the senses left for w are processed and the result is presented (step 5). To illustrate the process, consider the following text extracted from SemCor:

*The jury(2) praised the administration(3)*
*and operation(8) of the Atlanta*
*Police_Department(1), the*
*Fulton_Tax_Commissioner_'s_Office, the*
*Bellwood and Alpharetta prison_farms(1),*
*Grady_Hospital and the*
*Fulton_Health_Department.*

Figure 3: sample sentence from SemCor

The underlined words are nouns represented in WordNet with the number of senses between brackets. The noun to be disambiguated in our example is *operation.*, and a window size of five will be used.

**(step 1)** The following figure shows partially the lattice for the example sentence. As far as *Prison farm* appears in a different hierarchy we do not show it in figure 4:

```
police_department_0
   => local department, department of
   local government
      => government department
         => department
   jury_1, panel
      => committee, commission
   operation_3, function
      => division
         => administrative unit
            => unit
               => organization
                  => social group
                     => people
                        => group

administration_1, governance...
jury_2
   => body
      => people
         => group, grouping
```

Figure 4: partial lattice for the sample sentence

The concepts in WordNet are represented as lists of synonyms. Word senses to be disambiguated are shown in bold. Underlined concepts are those selected with highest Conceptual Density. Monosemic nouns have sense number 0.

**(Step 2)** `<administrative_unit>`, for instance, has underneath 3 senses to be disambiguated and a subhierarchy size of 96 and therefore gets a Conceptual Density of 0.256. Meanwhile, `<body>`, with 2 senses and subhierarchy size of 86, gets 0.062.

**(Step 3)** `<administrative_unit>`, being the concept with highest Conceptual Density is selected.

**(Step 4)** **operation_3**, **police_department_0** and **jury_1** are the senses chosen for *operation*, *Police Department* and *jury*. All the other concepts below `<administrative_unit>` are marked so that they are no longer selected. Other senses of those words are deleted from the lattice e.g. **jury_2**. In the next loop of the algorithm `<body>` will have only one disambiguation-word below it, and therefore its density will be much lower. At this point the algorithm detects that further disambiguation is not possible, and quits the loop.

**(Step 5)** The algorithm has disambiguated **operation_3**, **police_department_0**, **jury_1** and **prison_farm_0** (because this word is monosemous in WordNet), but the word *administration* is still ambiguous. The output of the algorithm , thus, will be that the sense for *operation* in this context, i.e. for this window, is **operation_3**. The disambiguation window will move rightwards, and the algorithm will try to disambiguate *Police Department* taking as context *administration*, *operation*, *prison farms* and whichever noun is first in the next sentence.

The disambiguation algorithm has and intermediate outcome between completely disambiguating a word or failing to do so. In some cases the algorithm returns several possible senses for a word. In this experiment we treat this cases as failure to disambiguate.

# 5 The Experiment

We selected one text from SemCor at random: br-a01 from the gender "Press: Reportage". This text is 2079 words long, and contains 564 nouns. Out of these, 100 were not found in WordNet. From the 464 nouns in WordNet, 149 are monosemic (32%).

The text plays both the role of input file (without semantic tags) and (tagged) test file. When it is treated as input file, we throw away all non-noun words, only leaving the lemmas of the nouns present in WordNet. The program does not face syntactic ambiguity, as the disambiguated part of speech information is in the input file. Multiple word entries are also available in the input file, as long as they are present in WordNet. Proper nouns have a similar treatment: we only consider those that can be found in WordNet. Figure 5 shows the way the algorithm would input the example sentence in figure 3 after stripping non-noun words.

After erasing the irrelevant information we get the words shown in figure 6[2].

The algorithm then produces a file with sense tags that can be compared automatically with the original file (c.f. figure 5).

```
<s>
<wd>jury</wd><sn>[noun.group.0]</sn><tag>NN</tag>
<wd>administration</wd><sn>[noun.act.0]</sn><tag>NN</tag>
<wd>operation</wd><sn>[noun.state.0]</sn><tag>NN</tag>
<wd>Police_Department</wd><sn>[noun.group.0]</sn><tag>NN</tag>
<wd>prison_farms</wd><mwd>prison_farm</mwd><msn>[noun.artifact.0]</msn><tag>NN</tag>
</s>
```

Figure 5: Semcor format

```
jury administration operation Police_Department prison_farm
```

Figure 6: input words

---

[2]*Note that we already have the knowledge that police department and prison farm are compound nouns, and that the lemma of prison farms is prison farm.*

Deciding the optimum context size for disambiguating using Conceptual Density is an important issue. One could assume that the more context there is, the better the disambiguation results would be. Our experiment shows that precision[3] increases for bigger windows, until it reaches window size 15, where it gets stabilised to start decreasing for sizes bigger than 25 (c.f. figure 7). Coverage over polysemous nouns behaves similarly, but with a more significant improvement. It tends to get its maximum over 80%, decreasing for window sizes bigger than 20.

Precision is given in terms of polysemous nouns only. The graphs are drawn against the size of the context[4] that was taken into account when disambiguating.

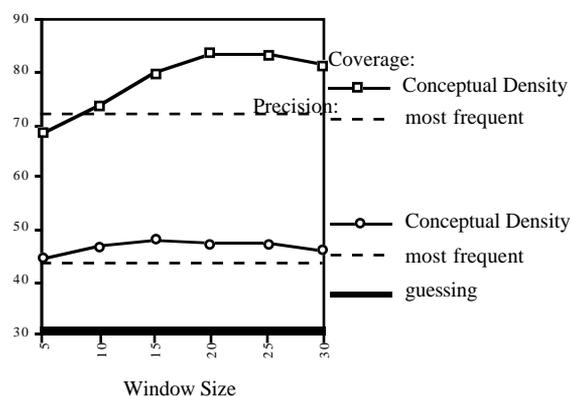

Figure 7: precision and coverage

Figure 7 also shows the guessing baseline, given when selecting senses at random. First, it was calculated analytically using the polysemy counts for the file, which gave 30% of precision. This result was checked experimentally running an algorithm ten times over the file, which confirmed the previous result.

We also compare the performance of our algorithm with that of the "most frequent" heuristic. The frequency counts for each sense were collected using the rest of SemCor, and then applied to the text. While the precision is similar to that of our algorithm, the coverage is nearly 10% worse.

All the data for the best window size can be seen in table 1. The precision and coverage shown in the preceding graph was for polysemous nouns only. If we also include monosemic nouns precision raises from 47.3% to 66.4%, and the coverage increases from 83.2% to 88.6%.

| % w=25 | Cover. | Prec. | Recall |
|---|---|---|---|
| polysemic | 83.2 | 47.3 | 39.4 |
| overall | 88.6 | 66.4 | 58.8 |

**Table 1:** overall data for the best window size

# 6 Further Work

Senses in WordNet are organised in lexicographic files which can be roughly taken also as a semantic classification. If the senses of a given word that are from the same lexicographic file were collapsed, we would disambiguate at a level closer to the homograph level of disambiguation.

Another possibility we are currently considering is the inclusion of meronymic relations in the Semantic Density algorithm. The more semantic information the algorithm gathers the better performance it can be expected.

At the moment of writing this paper more extensive experiments which include other three texts from SemCor are under way. With these experiments we would like to evaluate the two improvements outlined above. Moreover, we would like to check the performance of other algorithms for conceptual distance on the same set of texts.

This methodology has been also used for disambiguating nominal entries of bilingual MRDs against WordNet (Rigau & Agirre 95).

# 7 Conclusion

The automatic method for the disambiguation of nouns presented in this paper is ready-usable in any general domain and on free-running text, given part of speech tags. It does not need any training and uses word sense tags from WordNet, an extensively used lexical data base.

---

[3] *Precision is defined as the ratio between correctly disambiguated senses and total number of answered senses. Coverage is given by the ratio between total number of answered senses and total number of senses. Recall is defined as the ratio between correctly disambiguated senses and total number of senses.*

[4] *Context size is given in terms of nouns.*

The algorithm is theoretically motivated and founded, and offers a general measure of the semantic relatedness for any number of nouns in a text.

In the experiment, the algorithm disambiguated one text (2079 words long) of SemCor, a subset of the Brown corpus. The results were obtained automatically comparing the tags in SemCor with those computed by the algorithm, which would allow the comparison with other disambiguation methods.

The results are promising, considering the difficulty of the task (free running text, large number of senses per word in WordNet), and the lack of any discourse structure of the texts.

## Acknowledgements

We wish to thank all the staff at CRL in New Mexico State University, specially Jim Cowie, Joe Guthrie, Louise Guthrie and David Farwell. We would also like to thank Ander Murua, who provided mathematical assistance, Xabier Arregi, Jose Mari Arriola, Xabier Artola, Arantxa Diaz de Ilarraza, Kepa Sarasola, and Aitor Soroa from the Computer Science Department of EHU and Francesc Ribas, Horacio Rodríguez and Alicia Ageno from the CS Department of UPC.